\RequirePackage{fix-cm}
\documentclass[twocolumn]{svjour3}          
\smartqed  
\usepackage{graphicx}
\usepackage{cite}
\usepackage{amsmath,amssymb,amsfonts}
\usepackage{algorithmic}
\usepackage{textcomp}
\usepackage[table,usenames,dvipsnames]{xcolor}
\usepackage{pifont}
\usepackage{pgfplots}
\usepackage{url}
\usepackage{subcaption}
\newcommand{\figref}[1]{Fig.~\ref{#1}}
\newcommand{\secref}[1]{Section~\ref{#1}}
\newcommand{\tabref}[1]{Table~\ref{#1}}

\newcommand{\equref}[1]{Equation~\ref{#1}}

\definecolor{DarkPurple}{RGB}{17, 36, 107}
\definecolor{DarkBlue}{RGB}{15, 25, 76}
\definecolor{DarkBlue2}{RGB}{27, 98, 117}
\definecolor{DarkGreen}{RGB}{51, 76, 73}
\definecolor{DarkYellow}{RGB}{167, 172, 131}
\definecolor{Purple}{RGB}{132, 137, 185}
\definecolor{LightBlue}{RGB}{95, 173, 209}
\definecolor{LightGreen}{RGB}{168, 228, 170}
\definecolor{LightYellow}{RGB}{245, 236, 154}
\definecolor{Gray}{gray}{0.88}
\definecolor{DarkGray}{gray}{0.8}
\definecolor{LightGray}{gray}{0.93}
\definecolor{PastelPurple}{RGB}{97, 101, 136}
\definecolor{PastelBlue}{RGB}{95, 173, 209}
\definecolor{PastelGreen}{RGB}{90, 154, 90}

\pgfplotsset{compat=1.8}
\usepgfplotslibrary{statistics}

\newcommand{\greencheck}{{\color{ForestGreen}\checkmark}}
\newcommand{\redxmark}{{\color{Maroon}\ding{55}}}

\makeatletter
\newcommand*\bigcdot{\mathpalette\bigcdot@{.5}}
\newcommand*\bigcdot@[2]{\mathbin{\vcenter{\hbox{\scalebox{#2}{$\m@th#1\bullet$}}}}}
\makeatother
\journalname{Journal of Real-Time Image Processing}

\begin{document}

\title{Real-Time Online Unsupervised Domain Adaptation for Real-World Person Re-identification

\thanks{This research is supported by the National Science Foundation (NSF) under Award No. 1831795 and NSF Graduate Research Fellowship Award No. 1848727.}
}


\author{Christopher Neff         \and
        Armin Danesh~Pazho \and Hamed Tabkhi
}


\institute{Christopher Neff \at
              \email{cneff1@uncc.edu}           
           \and
           Armin Danesh~Pazho \at
              \email{adaneshp@uncc.edu}
           \and
           Hamed Tabkhi \at
              \email{htabkhiv@uncc.edu}
            \and University of North Carolina at Charlotte, NC, USA
}

\date{Received: date / Accepted: date}

\maketitle

\begin{abstract}
Following the popularity of Unsupervised Domain Adaptation (UDA) in person re-identification, the recently proposed setting of Online Unsupervised Domain Adaptation (OUDA) attempts to bridge the gap towards practical applications by introducing a consideration of streaming data. However, this still falls short of truly representing real-world applications. This paper defines the setting of Real-world Real-time Online Unsupervised Domain Adaptation (R$^2$OUDA) for Person Re-identification. The R$^2$OUDA setting sets the stage for true real-world real-time OUDA, bringing to light four major limitations found in real-world applications that are often neglected in current research: system generated person images, subset distribution selection, time-based data stream segmentation, and a segment-based time constraint. To address all aspects of this new R$^2$OUDA setting, this paper further proposes Real-World Real-Time Online Streaming Mutual Mean-Teaching (R$^2$MMT), a novel multi-camera system for real-world person re-identification. Taking a popular person re-identification dataset, R$^2$MMT was used to construct over 100 data subsets and train more than 3000 models, exploring the breadth of the R$^2$OUDA setting to understand the training time and accuracy trade-offs and limitations for real-world applications. R$^2$MMT, a real-world system able to respect the strict constraints of the proposed R$^2$OUDA setting, achieves accuracies within $0.1\%$ of comparable OUDA methods that cannot be applied directly to real-world applications.
\end{abstract}


\section{Introduction}
\label{sec:intro}

Person re-identification (ReID) is the task of matching a person in an image with other instances of that person in other images, either from the same camera or a different one. More specifically, it is associating a person's query with its match in a gallery of persons \cite{Survey2020}. Person ReID is a common task in many real-world applications. Such applications include video surveillance (\textit{e.g.} determining when unauthorized people are present in an area), public safety (\textit{e.g.} understanding pedestrian motion to avoid accidents), and smart health (\textit{e.g.} mobility assessment and fall detection for seniors needing assistance). Thus, achieving accurate and robust person ReID for any environment is an important research goal for the community.

\begin{table*}[ht]
    \centering
    \resizebox{0.95\linewidth}{!}{
    \begin{tabular}{c||c|c|c}
    Real-World & UDA & OUDA & \cellcolor{LightGray}R$^2$OUDA (Ours) \\ 
    \hline \hline
    Data from target domain is only available through a data stream. & \redxmark & \greencheck\textsuperscript{\dag} & \cellcolor{LightGray}\greencheck \\
    Person crops are not provided and must be generated online. & \redxmark & \redxmark & \cellcolor{LightGray}\greencheck \\
    There is no guarantee that every identity will be available during training. & \redxmark & \redxmark & \cellcolor{LightGray}\greencheck \\
    The distribution of person crops must be determined online. & \redxmark & \redxmark & \cellcolor{LightGray}\greencheck \\
    Training time must be accounted for.  & \redxmark & \redxmark & \cellcolor{LightGray}\greencheck \\

    \end{tabular}
    }
    \caption{Challenges of Real-World Applications and if they are addressed in the UDA, OUDA, and R$^2$OUDA settings. \\ 
    \textsuperscript{\dag} Streaming data is simulated.}
    \label{tab:setting}
\end{table*}

Many methods have been developed for person ReID \cite{ReIDWild,DefenseTripletLoss,CascadedPairwise,kreciprocal}, and many high quality datasets have been created for the task \cite{Market,CUHK03,PersonTransferGAN,DukeMTMC,MARS}. Deep learning approaches have been able to achieve incredible accuracies, nearly reaching saturation in some cases \cite{flipreid,Centroids,VALossReID,STReID}. However, person ReID is a highly context-specific task, and models trained on one dataset often fail to perform well on others \cite{Survey2020}.
Unsupervised Domain Adaptation (UDA) has been studied to combat this domain shift \cite{Survey2020,InstanceGuided,AdaptiveTransfer,PersonTransferGAN,ComplementaryPseudoLabels,TheoryPractice}. In UDA, initial training is performed on the labeled data of the source domain, and then inference is done in a different target domain. UDA methods generally achieve lower accuracies than State-of-the-Art (SotA) deep learning approaches that train directly on the target domain. However, recent approaches have begun to close that gap \cite{MMT,SpCL,HCR}.

One common thread among these approaches is the reliance on having the entirety of the target domain available at training time. While this is convenient for research, many practical applications do not have unrestricted access to the entire target domain. Recently, \cite{OUDA} introduced the setting of Online Unsupervised Domain Adaptation (OUDA). OUDA specifies that data from the target domain can only be accessed through a data stream, bringing research more in line with real-world applications. OUDA adopts a batch-based relaxation \cite{MemoryConstraints} where different identities are separated among batches to simulate streaming data. OUDA also argues that confidentiality regulations make it such that many real-world applications can only store data for a limited amount of time, applying a restriction that image data cannot be stored beyond the batch in which it was collected.

\tabref{tab:setting} shows the challenges of real-world applications, and how UDA and OUDA fail to fully address them. Like UDA before it, OUDA uses hand-crafted person ReID datasets for the target domain. Not only is the data stream only simulated, but the provided person images were hand selected by the creators of the dataset. In a real-world system, person images need to be generated by the system itself, creating a layer of noise not present in hand-crafted datasets. Further, by using hand-crafted datasets, the distribution of person images is guaranteed to be suitable for training. Specifically, most person ReID dataset tend to have a fairly uniform distribution, having around the same number of person images for each identity \cite{BottumUpClustering}. However, in real-world applications, there is no guarantee that person images generated from streaming data will form a uniform distribution in identities. There is also no guarantee that every identity in the dataset will be available for training.

To bring the field closer to the real-world, this paper proposes Real-World Real-Time Online Unsupervised Domain Adaptation (R$^2$OUDA), a setting designed to address the challenges found in real-world applications, as seen in \tabref{tab:setting}. R$^2$OUDA defines four major considerations beyond the OUDA setting needed to develop systems for the real world. First, R$^2$OUDA considers that person images must be generated algorithmically from streaming data. Second, the distribution of data to be used in training must also be determined algorithmically. Third, R$^2$OUDA expands the batched-based relaxation \cite{MemoryConstraints} of online learning to use time segments, relating the conceptual mini-batch to the real-world notion of time inherent in streaming data. Fourth, R$^2$OUDA defines a time constraint such that the time spent training a single time segment cannot interfere with the training for subsequent time segments.

To address all aspects of the new R$^2$OUDA setting, this paper further proposes Real-World Real-Time Online Streaming Mutual Mean-Teaching (R$^2$MMT). R$^2$MMT is an end-to-end multi-camera system designed for real-world person ReID. Using object detection, pedestrian tracking, human pose estimation, and a novel approach for Subset Distribution Selection (SDS), R$^2$MMT is able to generate person crops directly from a data stream, filter them based on representation quality, and create a subset for training with a suitable distribution. To show the viability of R$^2$MMT to meet the challenges of real-world applications, and to explore the breadth of the R$^2$OUDA setting, an exhaustive set of experiments were conducted on the popular and challenging DukeMTMC dataset \cite{DukeMTMC}. Using R$^2$MMT, over 100 data subsets were created and more than 3000 models were trained, capturing the trade-offs and limitations of real-world applications and the R$^2$OUDA setting. R$^2$MMT is a real-world system that can meet the demanding requirements of the proposed R$^2$OUDA setting, and is able to achieve over $73\%$ Top-1 accuracy on DukeMTMC-reid, within $0.1\%$ of comparable OUDA methods that cannot be directly applied for real-world applications. 

To summarize, this paper's contributions are as follows:
\begin{itemize}
    \item We define the setting of Real-World Real-Time Online Unsupervised Domain Adaptation, accounting for the challenges of real-world applications and bridging the gap between research and application.
    \item We propose Real-World Real-Time Online Streaming Mutal Mean-Teaching, a novel end-to-end multi-camera person ReID system designed to meet the challenges of R$^2$OUDA and real-world applications.
    \item We perform exhaustive experimentation, creating over 100 data subsets and training over 3000 models, to explore the breadth of the R$^2$OUDA setting and understand the trade-offs and limitations of real-world applications.
\end{itemize}

\section{Related Work} \label{sec:relatedwork}

The UDA setting for person ReID has been extensively explored by the research community \cite{Survey2020,OpenWorldSurvey,PastPresentFuture,TheoryPractice}. In general, there are two main categories of algorithms used to perform UDA for person ReID: style transfer methods and target domain clustering methods.

\subsection{Style Transfer}

Style transfer based methods generally use Generative Adversarial Networks (GANs) \cite{GANs} to perform image-to-image translation \cite{Pix2Pix}, modifying images from the source domain to look like the target domain without affecting the context of the original images. \cite{PreservedSelf} uses self-similarity and domain-dissimilarity to ensure transferred images maintain cues to the original identity without matching to other identities in the target domain, while \cite{RelationRegularization} introduces an online relation-consistency regularization term to ensure relations of the source domain are kept after transfer to the target domain. \cite{AdaptiveTransfer} separates transfers into factor-wise sub-transfers, across illumination, resolution, and camera view, to better fit the source images into the target domain. \cite{InstanceGuided} uses a dual conditional GAN to transfer source domain images to multiple styles in the target domain, creating a multitude of training instances for each source identity. \cite{PersonTransferGAN} uses a cycle consistent loss \cite{UnpairedI2I} with an emphasis on the foreground to better maintain identities between styles. \cite{SBSGAN} looks at domain shift as background shift and uses a GAN to remove backgrounds without damaging foregrounds, while a densely associated 2-stream network integrates identity related cues present in backgrounds.

\subsection{Target Domain Clustering}

Target domain clustering approaches focus on using clustering algorithms to group features of the target domain for use as labels to fine tune a neural network pre-trained on the source domain \cite{ClusteringFineTuning}. This is usually done in an iterative fashion, where clustering is performed between training epochs to update the group labels as the model learns. \cite{DynamicGraph} proposes using a dynamic graph matching framework to better handle large cross-camera variations. \cite{SelfSimilarityGrouping} introduces a self-similarity group to leverage part-based similarity to build clusters from different camera views. \cite{BottumUpClustering} utilizes a diversity regularization term to enforce a uniform distribution among the sizes of clusters. \cite{SelfPacedContrast} introduces hybrid memory to dynamically generate instance-level supervisory signal for feature representation learning. \cite{MMT} builds on \cite{MeanTeachers}, using two teacher models and their temporally averaged weights to produce soft pseudo labels for target domain clustering. \cite{CANUReID} utilizes both target domain clustering and adversarial learning to create camera invariant features and improve target domain feature learning. 

\subsection{Online Unsupervised Domain Adaptation}

While Online Unsupervised Domain Adaptation has been explored for other AI tasks \cite{OUDACovid19,OUDADiscrepancies,OUDAECG,OUDAMultistage,OUDAMultistep,OUDASeg,OUDASegFrequency}, it was first defined for the field of person ReID in \cite{OUDA}. OUDA for Person ReID aims to create a practical online setting similar to that found in practical applications. OUDA builds upon the UDA setting by adding two considerations. First, data from the target domain is accessed via a data stream and not available all at once. Second, due to confidentiality concerns common in many countries, data from the target domain can only be stored for a limited time and only model parameters trained on that data may be persistent.


\section{Proposed R$^2$OUDA Setting}\label{sec:OUDA}

\begin{figure*}[ht!]
    \centering
    \resizebox{0.95\linewidth}{!}{
    \includegraphics[clip,trim={21 18 20 21},width=0.95\textwidth]{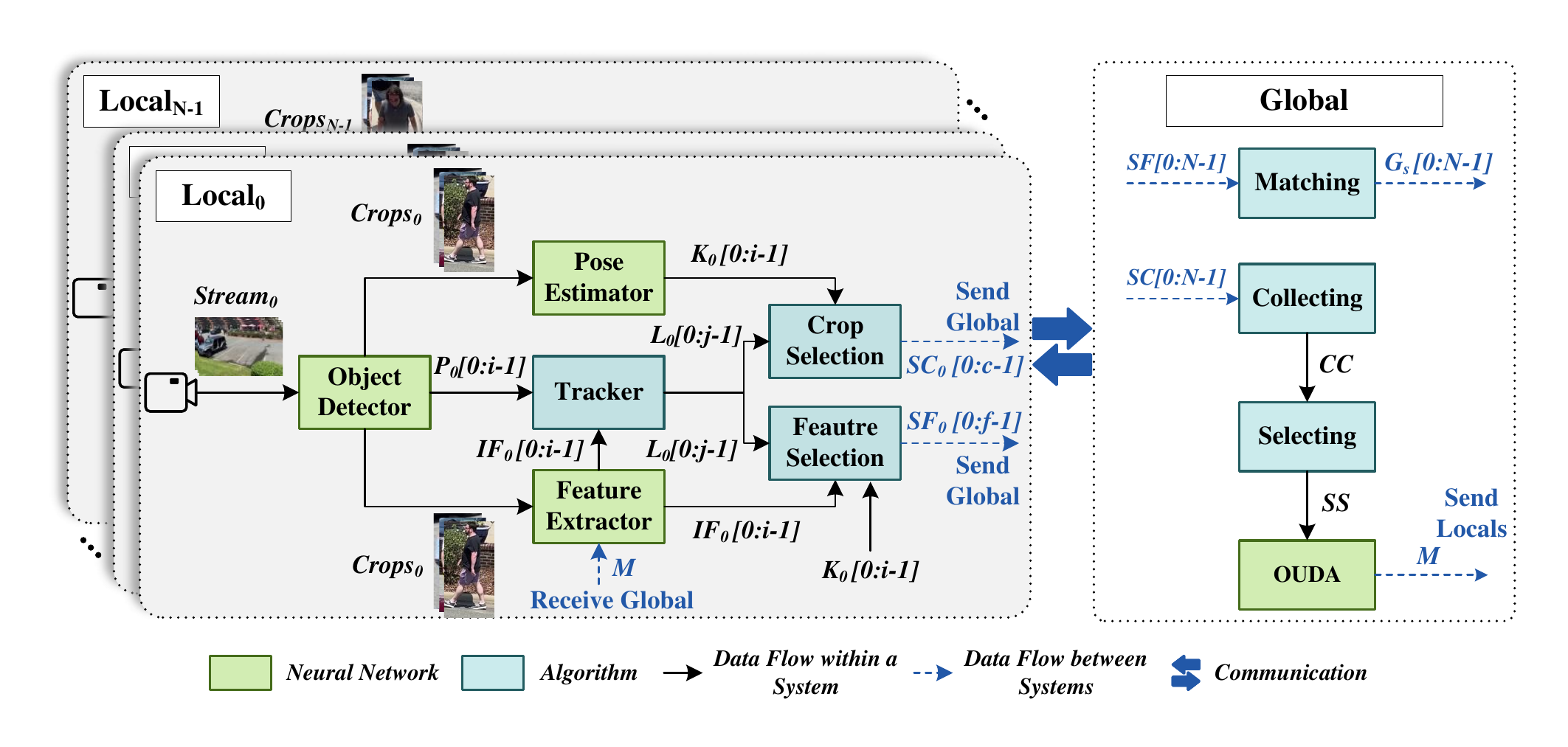}
    }
    \caption{System view of Real-World Real-Time Online Streaming Mutual Mean-Teaching.}
    \label{fig:system}
\end{figure*}

The proposed setting of Real-World Real-Time Online Unsupervised Domain Adaptation, building off OUDA \cite{OUDA}, considers that we have access to a completely annotated source dataset $D_{S}$ as well as partial access to an unlabeled target dataset $D_{T}$ in the domain of our target application. In contrast to standard UDA, in both OUDA and R$^2$OUDA the data from $D_{T}$ is only accessible as an online stream of data. Whereas both UDA and OUDA use person crops from hand crafted datasets, R$^2$OUDA specifies that person crops from $D_{T}$ must be generated algorithmically from the data stream. This reflects how data is gathered in the real world. Where hand selected crops from datasets are generally highly representative, crops generated from a data stream will have varying levels of quality. This introduces noise in $D_{T}$, both in quality and in the inevitable missed detections, which needs to be accounted for.

Additionally, hand crafted datasets choose person images to fit a distribution suitable for training. However, since crops in R$^2$OUDA are generated from streaming data, such a distribution can not be assumed. This leads to the second consideration of R$^2$OUDA, that the distribution of data to be used in training must be determined algorithmically. Instead of relying on a predefined set of person images, systems must generate their own data subset, determining its size and distribution appropriately. This also reflects the real-world, as it is rarely known beforehand the amount and distribution of person crops that will be collected by an application.

Continuing with the batched-based relaxation \cite{MemoryConstraints} of the online learning scenario proposed in \cite{OUDA}, we further introduce a time constraint for R$^2$OUDA. First, instead of separating our "mini-batches" ("tasks" as defined in \cite{OUDA}) across identities, since R$^2$OUDA requires actual streaming data, the data stream is separated into discrete time segments. We consider that for a chosen time segment of length $\tau$, the streaming data will be divided into equal, non-overlapping time segments of length $\tau$ whose combined contents are equivalent to the original data stream.

For R$^2$OUDA, we must account both for applications that run continuously (i.e. the total length of the data stream is infinite) and the fact that, in the real world, computation resources are not unlimited. This leads to the necessity of a time constraint, but one that is not simple to define. Training time is inherently linked to hardware, and there are many techniques to hide latency or increase throughput in system design. As such, we simply define the time constraint such that, for any time segment $\tau_{i}$, the length of time spent training on data collected during $\tau_{i}$ must be such to not interfere with the training for the data collected during $\tau_{i+1}$. This is to prevent the training time deficit from increasing infinitely as $i$ increases.

In summary, R$^2$OUDA introduces four new considerations to better match real-world applications:
\begin{itemize}
    \item Person crops from the target domain must be generated algorithmically from a data stream.
    \item The selection and distribution of data to be used in training must be determined algorithmically.
    \item An expansion of the batch-based relaxation to use time segments, relating the conceptual mini-batch to the real-world notion of time inherent in streaming data.
    \item An additional time constraint such that the time spent training a single time segment cannot interfere with the training for any subsequent time segments.
\end{itemize}


\section{Real-World Real-Time Online Streaming MMT}
\label{sec:method}

\begin{figure*}[t!]
    \centering
    \resizebox{0.95\linewidth}{!}{
    \includegraphics[clip,trim={25 22 25 18},width=0.95\textwidth]{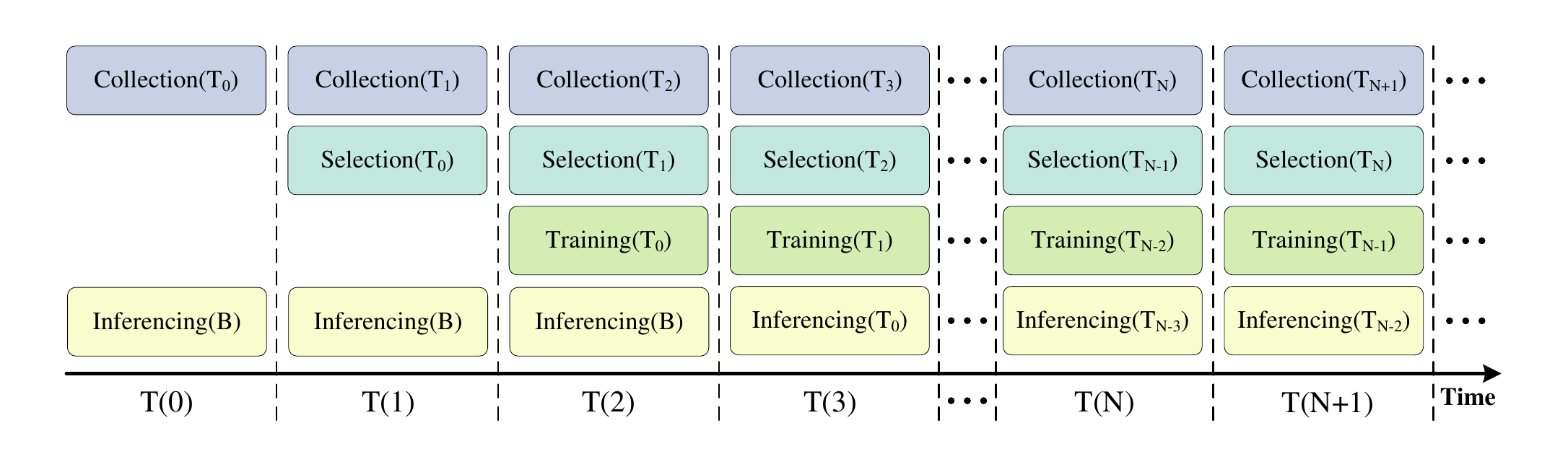}
    }
    \caption{Illustration of computation overlap through time.}
    \label{fig:pipeline}
\end{figure*}

To address the challenges of R$^2$OUDA, we present Real-World Real-Time Online Streaming Mutual Mean-Teaching, a novel multi-camera system for real-world person ReID. Similar to \cite{REVAMP2T}, R$^2$MMT is comprised of multiple Local Nodes and a single Global Node. Local nodes have access to the data stream directly from the cameras and are responsible for generating quality person images. The Global Node has access to all data generated by Local Nodes and is responsible for global ReID, subset distribution selection, and target domain training. An overview of R$^2$MMT can be seen in \figref{fig:system}.

On the Local Node, YOLOv5 \cite{YOLOv5} is used as an object detector to find people in the video stream. Image crops are created for each person and sent to both a pose estimator (HRNet \cite{HRNet}) and a ReID feature extractor (ResNet-50 \cite{ResNet}). Coordinates for each person and features generated by the feature extractor are sent to a tracker \cite{DeepSORT} for local ReID. Afterward, feature and crop selection are performed to ensure that features and person crops sent to the Global Node for global ReID and crop collection are highly representative. This process utilizes person bounding box coordinates from the tracker to filter out any persons that have significant overlap (IoU $>= 0.3$) with other persons. This limits the number of crops used for training and features used for ReID contain multiple persons. The pose estimator is used to determine the quality of the features themselves. We reason that if a highly representative feature is present, then poses generated from the person crop should be of high confidence, while the number of keypoints present can help determine if there is significant occlusion or cutoff. Only crops and features with poses containing 15 or more keypoints (out of 17 total \cite{COCO}) with at least $50\%$ confidence are sent to the Global Node.

On the Global Node, local identities and features are received from the Local Nodes and sent to a matching algorithm. This matching algorithm, as described in \cite{REVAMP2T}, performs global (i.e. multi-camera) ReID. Concurrently, person crops from all cameras are collected for a single time segment. Generally, far more features will be collected than can reasonably be used during training. For instance, when DukeMTMC-Video \cite{DukeMTMC} is sampled every frame, the system produces over 4 million crops that pass feature selection. To reduce redundancy and computation, R$^2$MMT samples crops for selection once every 60 frames.

After all person crops from a single time segment are collected, the Subset Distribution Selection algorithm is used to create a subset that maintains a distribution and number of crops suitable for training. R$^2$MMT uses an SDS algorithm based on the metric facility location problem \cite{HeuristicKCenters}. We define that given a number of features in a metric space, we wish to find a subset of $k$ features such that the minimum distance between any two features within the subset is maximized. However, this problem is known to be NP-hard \cite{GreedyKCenters}, making it unsuitable for our real-world applications. R$^2$MMT instead uses a greedy implementation of the algorithm proven to be $\Omega(log\:k)$-competitive with the optimal solution while proving to be significantly faster, especially for larger sets of data \cite{kmeans++}. For ease of readability, we adopt the nomenclature of $K$ to mean the number of instances per identity. Therefore the total number of person crops in a subset $k$ is equal to the number of identities in the dataset times $K$. To further reduce complexity, SDS is performed on the data from each camera individually, and their results are combined to form the complete subset.

Once the training subset is created, domain adaptation is performed using Mutual-Mean Teaching (MMT) \cite{MMT}. R$^2$MMT follows the training methodology described in \cite{MMT}, except that epochs and iterations are variable. Clustering is done using DBSCAN \cite{DBSCAN}, as GPU acceleration allows it to perform much faster than CPU based approaches. Exact training parameters, both for pre-training on the source domain and domain transfer on the target domain, are as detailed in \cite{MMT} unless otherwise noted.

Both SDS and training are time consuming, particularly when dealing with large amounts of data. To meet the time constraint of the R$^2$OUDA setting, R$^2$MMT utilizes a pipelined processing model, taking advantage of parallel computing resources while hiding the latency of the aforementioned tasks. An illustration of this pipelined approach can be seen in \figref{fig:pipeline}. Crop collection, SDS, and training are separated into their own pipeline stages. This means that while a model collects data for the current time segment, SDS on that data will occur the following time segment, and the training for that subset will occur the time segment after that. More formally, during a single time segment $T_N$, a model trained on data from $T_{N-3}$ is used to collect data from time segment $T_{N}$, while subset distribution selection is performed on data collected during $T_{N-1}$ and another network is being trained on a subset created from data from $T_{N-2}$. All of these processes will finish before $T_{N+1}$. This means there will always be a latency of two time segments between collection and inference for a single time segment. However, due to the pipeline structure, training throughput remains at a rate of one time segment per time segment. This satisfies the time constraint of R$^2$OUDA. 


\section{Experimental Results}
\label{sec:results}

\begin{figure*}
    \centering
    \begin{subfigure}[b]{0.3\textwidth}
        \centering
        \includegraphics[clip,trim={10 24 25 20}, width=\textwidth]{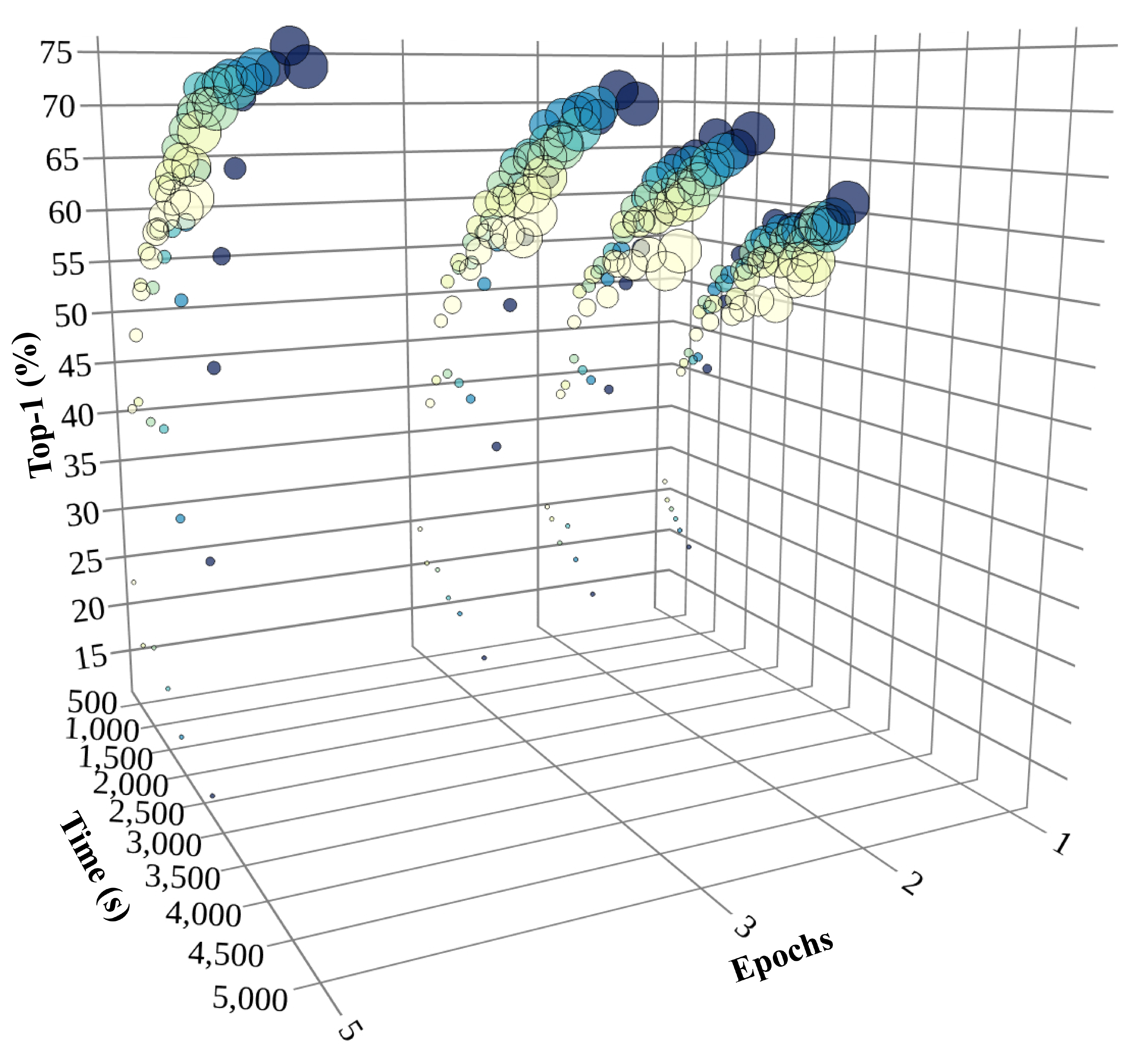}
        \caption{}
        \label{fig:phase1-1}
    \end{subfigure}
    \hfill
    \begin{subfigure}[b]{0.3\textwidth}
        \centering
        \includegraphics[clip,trim={5 23 10 40}, width=\textwidth]{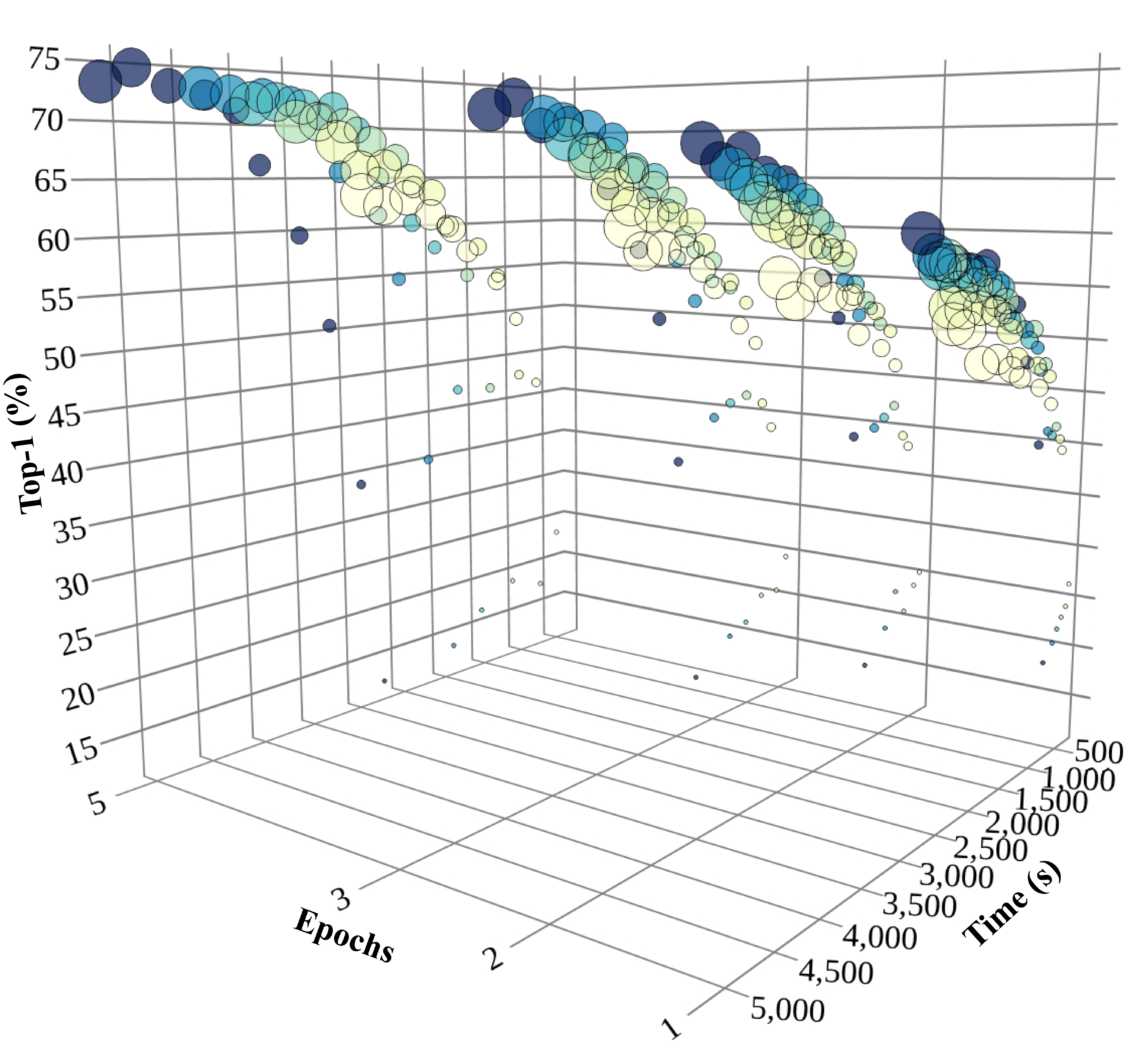}
        \caption{}
        \label{fig:phase1-2}
    \end{subfigure}
    \hfill
    \begin{subfigure}[b]{0.3\textwidth}
        \centering
        \includegraphics[width=\textwidth, trim={10 5 20 10},clip]{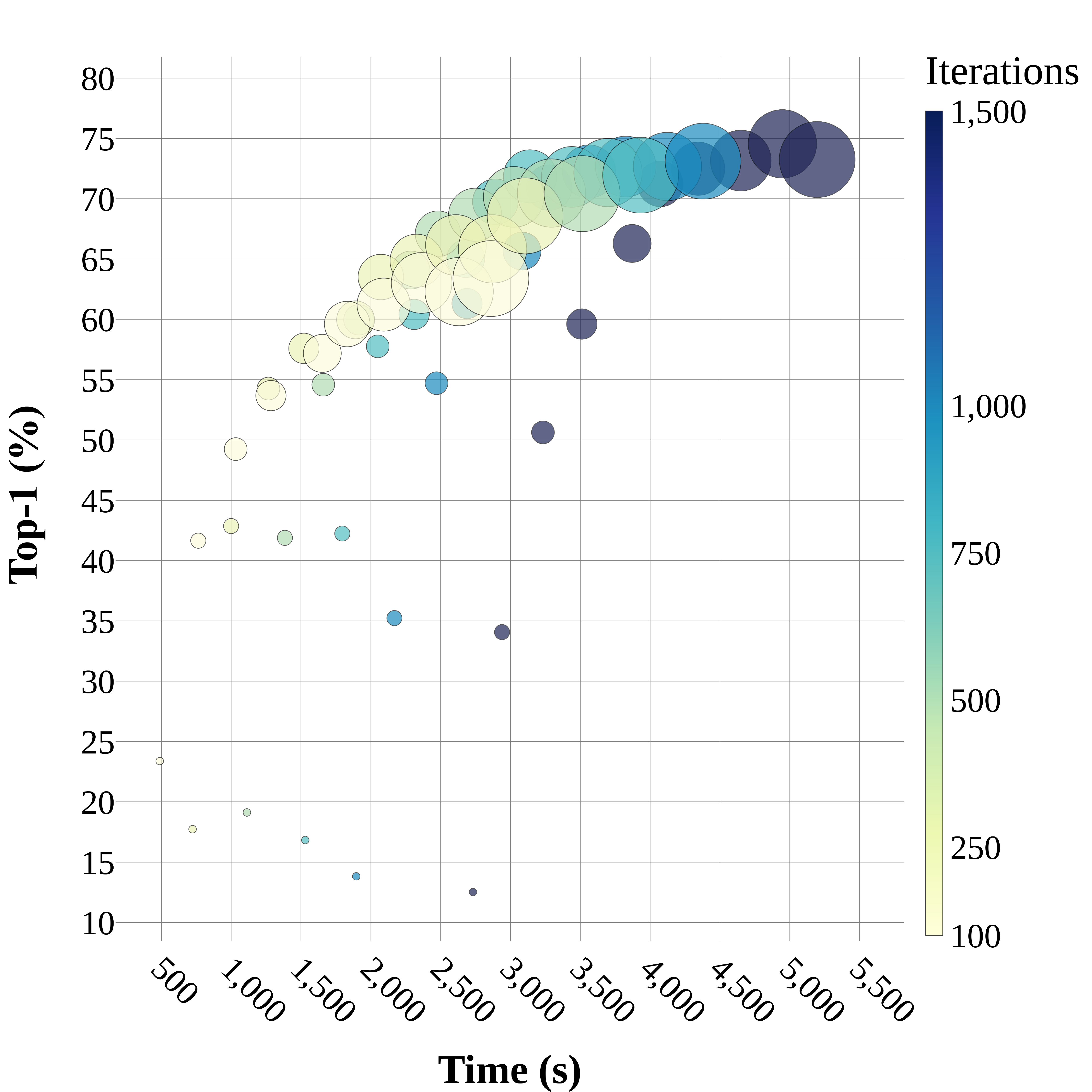}
        \caption{}
        \label{fig:pahse1-3}
    \end{subfigure}
    \caption{Results exploring SDS on the hand crafted DukeMTMC-reid dataset \cite{DukeMTMC}. (a) and (b) show two views of the results plotted in three-dimensional space, while (c) shows a two-dimensional view when $E=5$. Larger circles represent larger values of $k$.}
    \label{fig:phase1}
\end{figure*}

To explore the setting of R$^2$OUDA, we select the Market 1501 dataset \cite{Market} as the source domain and the DukeMTMC dataset \cite{DukeMTMC} as the target domain. The DukeMTMC dataset is desirable as a target domain because it has both a video dataset (DukeMTMC-video) and a hand crafted person ReID dataset (DukeMTMC-reid), both in the same domain. The video dataset is required in order to satisfy the streaming data constraint of the R$^2$OUDA setting. The hand crafted ReID dataset brings two benefits. First, it allows us to directly observe the effect of noisy system generated crops compared hand selected person images when used for training. Second, testing on the ReID dataset allows direct comparison with works done in the UDA and OUDA space. As such, all our Top-1 accuracies are reported on the DukeMTMC-reid dataset. Similarly, we determining subset size, we treat the number of identities for both DukeMTMC-reid and DukeMTMC-video to be 702, as described in \cite{DukeMTMC}. The number of person crops in a subset $k$ is always equal to $k\times702$. 

For all experiments, R$^2$MMT is used to perform domain adaptation. Parameters in all experiments are the same as in \cite{MMT}, except where noted otherwise. All Local Nodes are run on a single server with two AMD EPYC 7513 CPUs, 256 GB of RAM, and three Nvidia V100 GPUs. The Global Node is run on a workstation with an AMD Threadripper Pro 3975WX CPU, 256 GB RAM, and three Nvidia RTX A6000 GPUs. All timing results presented in this section are using this Global Node.

\subsection{Subset Distribution Selection} \label{sec:phase1}

We first explore the effect of using our baseline Subset Distribution Selection algorithm for training on the DukeMTMC-reid dataset. By using hand selected person crops from the dataset, we remove the effect of noise generated by our system and single out the impact of our SDS algorithm and the reduction in amount of data on domain adaptation. We vary the number of person images per identity $K$, iterations per epoch $I$, and total epochs $E$ as shown below. Note that using the entire DukeMTMC-reid dataset would be equivalent to $K=25$.

\begin{equation}
\begin{split}
    K & \in [2, 4, 6, 8, 10, 12, 14, 16, 18, 20] \\
    I & \in [100, 250, 500, 750, 1000, 1500] \\
    E & \in [1, 2, 3, 5]
\end{split}
\end{equation}

These variable ranges lead to 240 training permutations, which is difficult to list in a single table. Instead, the results are plotted in a three-dimensional space and can be seen in \figref{fig:phase1}. \textit{Training Time} and \textit{Top-1} make up the x and y axes, \textit{Epochs} are the z axis, \textit{Iterations} are noted by color, and $k$ is indicated by size, with bigger circles representing higher values of $k$. As the purpose of these experiments is to focus on the effects of our SDS algorithm, the system pipeline described in \secref{sec:method} is ignored and timing results count SDS and training sequentially. More detailed information on these experiments can be found in the supplementary materials.

From these graphs, we can understand the general trend of the data. Intuitively, we see a fairly linear trend where more data generally results in higher Top-1 accuracy. Likewise, more iterations per epoch and more epochs also tend to result in higher accuracy. Interestingly, with lower values of $k$ we see the reverse effect; more time spent training results in decreased accuracy, sometimes even below the pre-trained accuracy of 42.0\%. In general, at least 6 person images per identity are needed to consistently learn, while we start to see diminishing returns at around 16 person images per identity. The top result occurs when $K=20$, $I=1500$, and $E=5$, achieving a Top-1 accuracy of $74.55\%$ with a training time of $82$ minutes. This is only $3.5\%$ less than what comparable algorithms are able to achieve in the UDA setting \cite{MMT} and over $2\%$ greater than the same algorithm in the OUDA setting \cite{OUDA}. When using the same hardware, R$^2$MMT is $2.6\times$ faster than its UDA counterpart.

\subsection{System Generated Data} \label{sec:phase2}

\begin{figure*}
    \centering
    \begin{subfigure}[b]{0.3\textwidth}
        \centering
        \includegraphics[width=\textwidth]{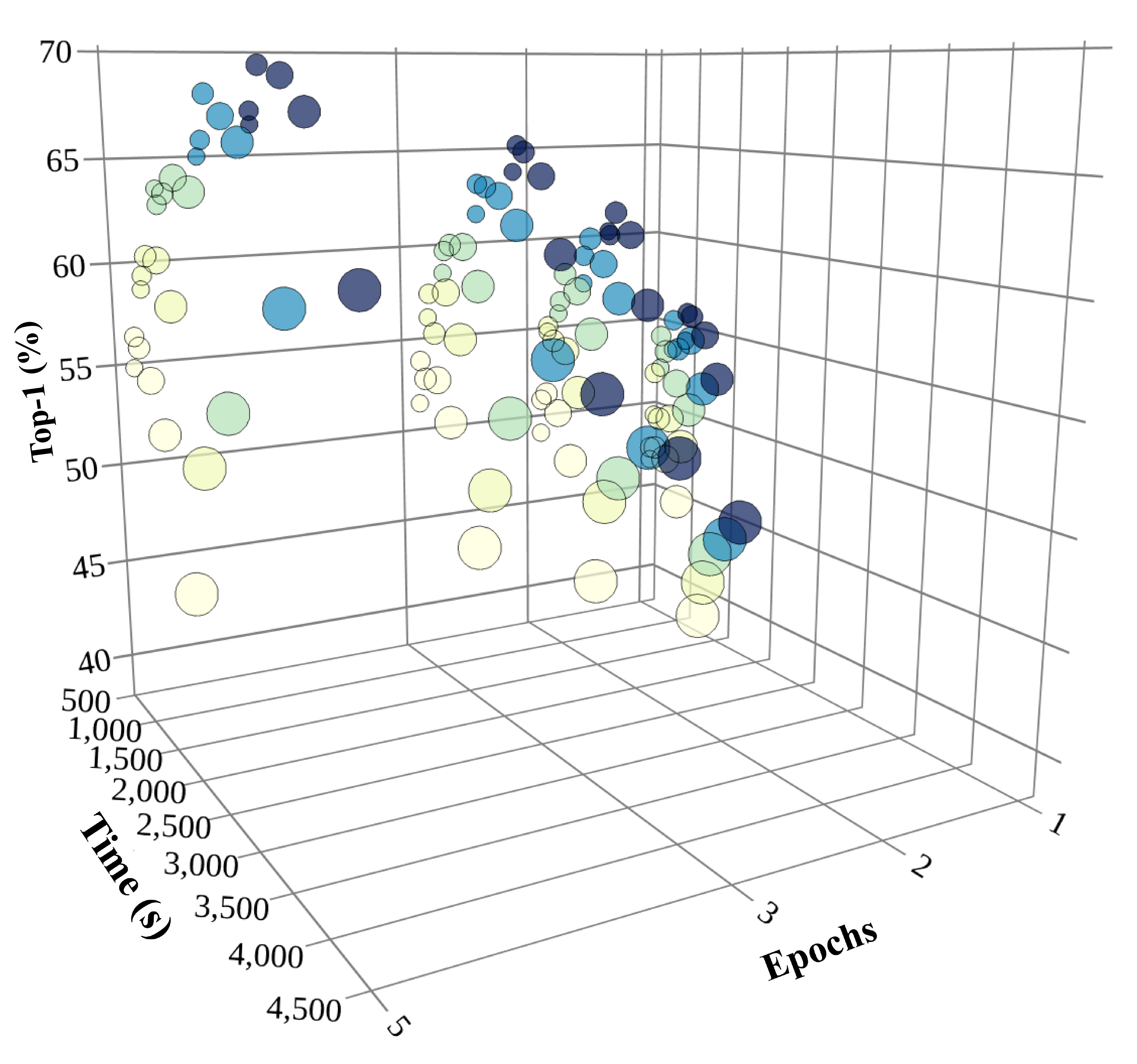}
        \caption{}
        \label{fig:phase2-1}
    \end{subfigure}
    \hfill
    \begin{subfigure}[b]{0.3\textwidth}
        \centering
        \includegraphics[width=\textwidth]{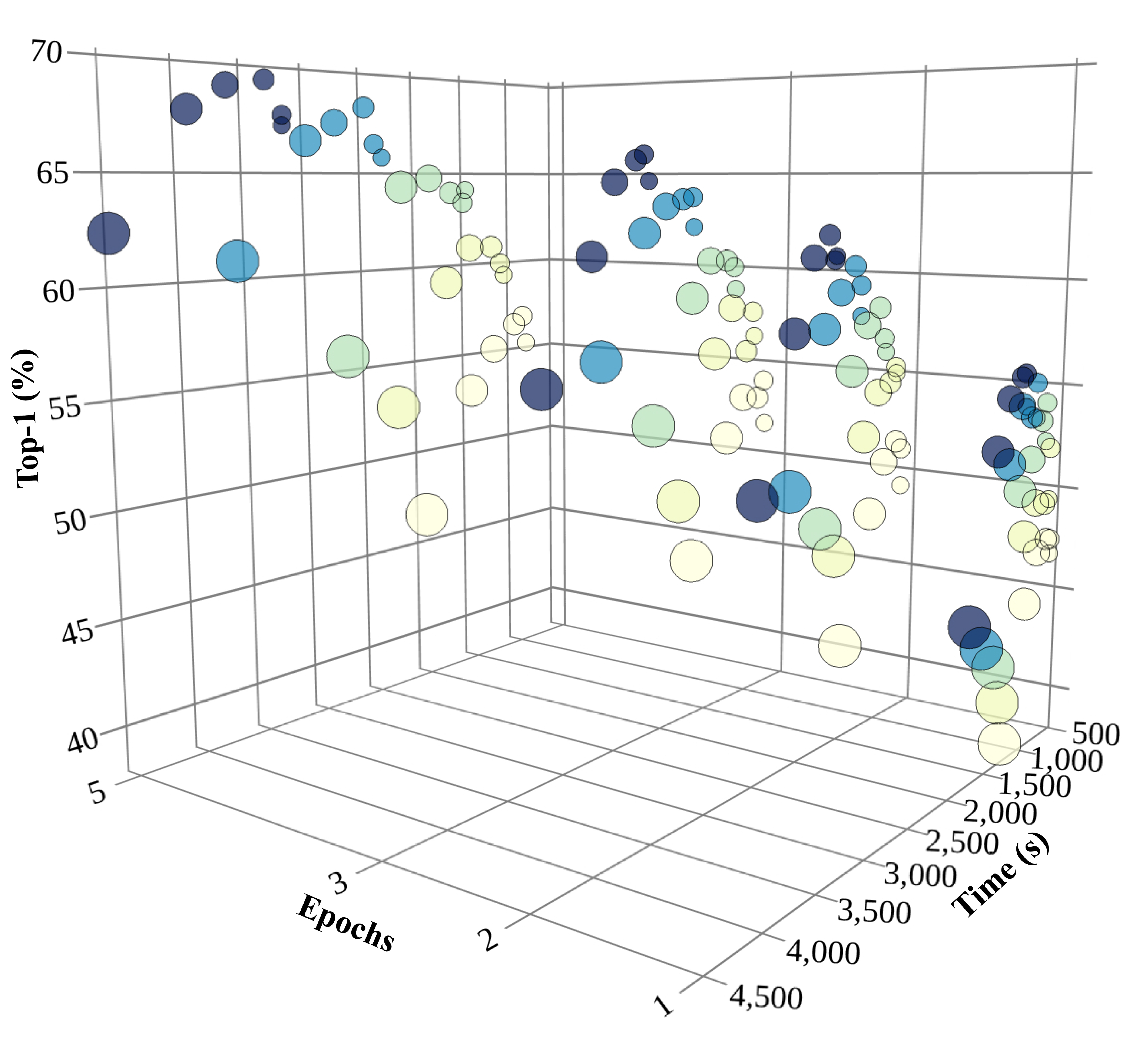}
        \caption{}
        \label{fig:phase2-2}
    \end{subfigure}
    \hfill
    \begin{subfigure}[b]{0.3\textwidth}
        \centering
        \includegraphics[width=\textwidth, trim={10 5 20 10},clip]{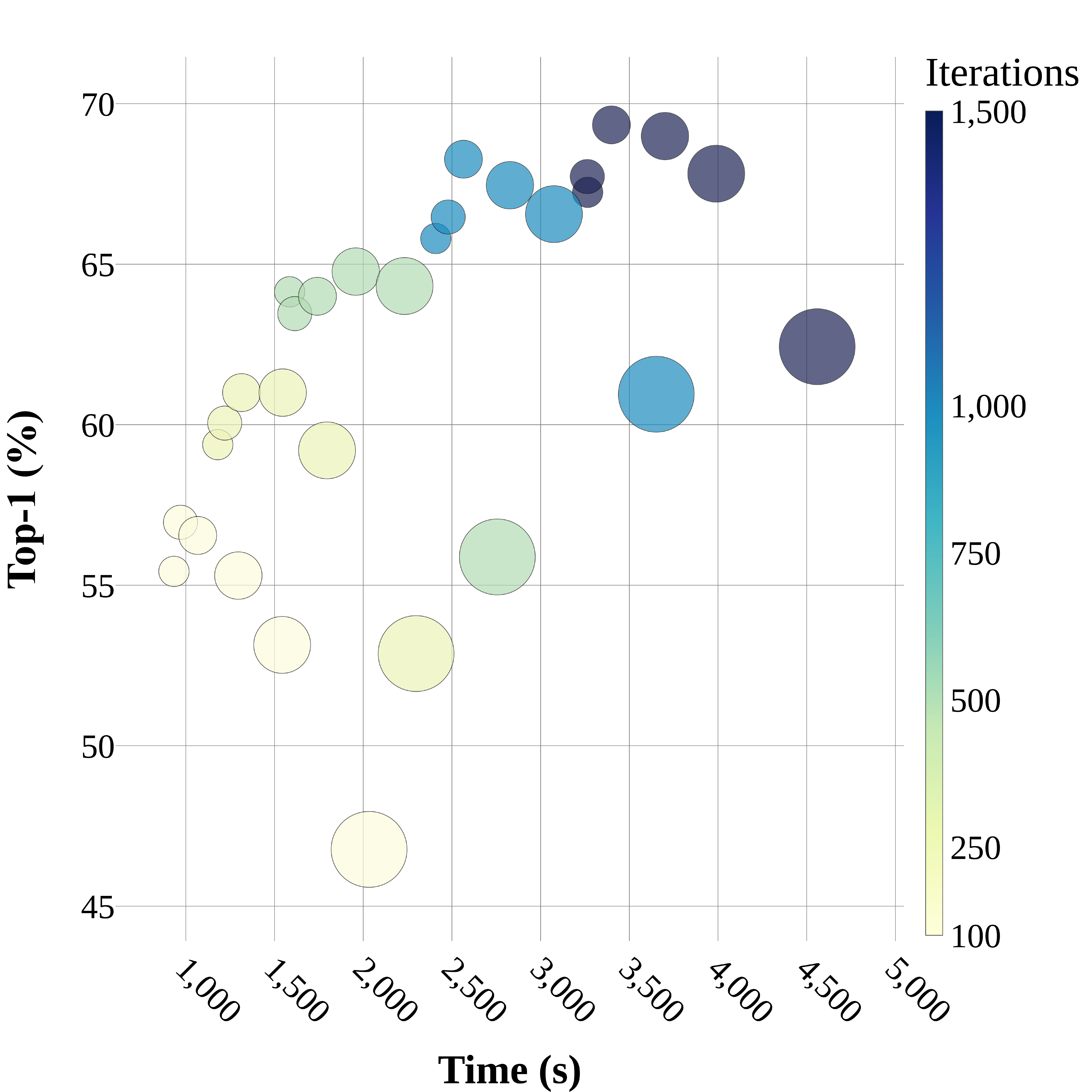}
        \caption{}
        \label{fig:phase2-3}
    \end{subfigure}
    \caption{Results exploring the use of system generated data using DukeMTMC-video \cite{DukeMTMC}. (a) and (b) show two views of the results plotted in three-dimensional space, while (c) shows a two-dimensional view when $E=5$. Larger circles represent larger values of $k$.}
    \label{fig:phase2}
\end{figure*}

As explained in \secref{sec:OUDA}, one of the requirements of the R$^2$OUDA setting is that person crops must be generated algorithmically from a data stream. As such, it is necessary to explore the effects of the noise this introduces. The structure of these experiments are exactly the same as in \secref{sec:phase1}, except that instead of using DukeMTMC-reid, R$^2$MMT generates data from the DukeMTMC-video dataset. Similar to \secref{sec:phase1}, we ignore the system pipeline and focus on the effects of the generated data. Based on the larger amount of data available in DukeMTMC-video, the ranges for our experimental variables are adjusted as shown below. Using all generated data would be equivalent to $K=99$.

\begin{equation}
\begin{split}
    K & \in [16, 18, 20, 25, 30, 40] \\
    I & \in [100, 250, 500, 1000, 1500] \\
    E & \in [1, 2, 3, 5]
\end{split}
\end{equation}

The results of this exploration can be seen in \figref{fig:phase2}, with more details available in the supplementary materials. Axes are identical to \figref{fig:phase1}, with color and size representing iterations and $k$ respectively. These graphs show a somewhat similar trend as in \secref{sec:phase1} with some interesting deviations. While the trend starts off with accuracy increasing as $k$ gets larger, there is a sharp decrease in accuracy when $k$ increases beyond a certain point. The scale of the decrease, as well as how early it occurs, lessens with both iterations and epochs. This is likely a byproduct of how many identities are present in DukeMTMC-video. While DukeMTMC only labels a total of 1404 identities, our system is able to detect far more. Increasing iterations has such a drastic effect here because it determines how many of and how often these identities are seen during an epoch. Further increasing iterations and epochs could help mitigate this, but would also increase overall training time. This, combined with the fact that more epochs and more iterations always result in higher accuracy, suggests that accuracy saturation has not been reached here, and the main limiting factor is training time. The highest accuracy achieved on this noisy data was a Top-1 of $69.34\%$, with $K=20$, $I=1500$, $E=5$, and a total training time of just under $57$ minutes. This is notably worse than both the $74.55\%$ achieved in \secref{sec:phase1} and the $72.3\%$ MMT achieves in the OUDA setting \cite{OUDA}. This demonstrates the extreme impact noisy data can have on unsupervised domain adaptation, and why the extra considerations of the R$^2$OUDA setting are a necessity when designing algorithms for real-world applications.

\subsection{R$^2$MMT}\label{sec:phase3}

Finally, we make the first attempt at addressing the R$^2$OUDA setting. An exhaustive set of experiments are conducted with R$^2$MMT, producing a fully functional, end-to-end system that meets all the requirements of the R$^2$OUDA setting. R$^2$MMT generates person crops from a stream of data, uses SDS to construct training subsets, operates on the notion of time segments, and must adhere to the strict time constraint outlined in \secref{sec:OUDA}. A successful implementation will conform to all of those standards while achieving the highest accuracy possible, ideally within range of what was seen in \secref{sec:phase1}.

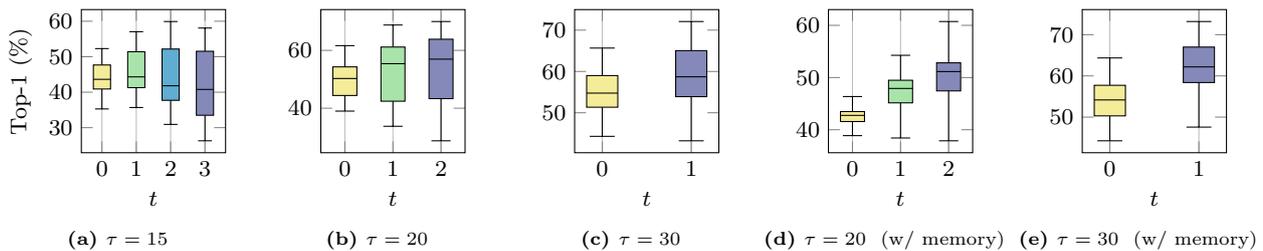
\begin{figure*}[!ht]
    \centering
    \begin{subfigure}[b]{0.2\textwidth}
        \centering
\begin{tikzpicture}
  \begin{axis}
    [
    boxplot/draw direction=y,
    xmajorgrids=true,
    width= \linewidth,
    trim left = 1cm,
    xtick={1, 2, 3, 4},
    xlabel = $t$,
    ylabel = Top-1 (\%),
    xticklabels={0, 1, 2, 3},
    boxplot/box extend=0.50,
    height = 1.0\linewidth,
    ]
    \addplot+[
    boxplot prepared={
      median=43.63,
      upper quartile=47.67,
      lower quartile=40.90,
      upper whisker=52.29,
      lower whisker=35.28
    }, draw=black, fill=LightYellow,
    ] coordinates {};
    \addplot+[
    boxplot prepared={
      median=44.32,
      upper quartile=51.40,
      lower quartile=41.28,
      upper whisker=57.05,
      lower whisker=35.68
    }, draw=black, fill=LightGreen,
    ] coordinates {};
    \addplot+[
    boxplot prepared={
      median=41.81,
      upper quartile=52.17,
      lower quartile=37.71,
      upper whisker=59.92,
      lower whisker=30.92
    }, draw=black, fill=LightBlue,
    ] coordinates {};
    \addplot+[
    boxplot prepared={
      median=40.78,
      upper quartile=51.54,
      lower quartile=33.50,
      upper whisker=58.08,
      lower whisker=26.30
    }, draw=black, fill=Purple,
    ] coordinates {};
  \end{axis}
\end{tikzpicture}
        \caption{$\tau=15$}
        \label{fig:bw15s}
    \end{subfigure}
    \hspace{-8pt}
    \begin{subfigure}[b]{0.2\textwidth}
        \centering
\begin{tikzpicture}
  \begin{axis}
    [
    boxplot/draw direction=y,
    xmajorgrids=true,
    width= \linewidth,
    trim left = 10cm,
    xtick={1, 2, 3},
    xlabel = $t$,
    xticklabels={0, 1, 2},
    boxplot/box extend=0.5,
    height = 1.0\linewidth,
    ]
    \addplot+[
    boxplot prepared={
      median=50.27,
      upper quartile=54.29,
      lower quartile=44.39,
      upper whisker=61.63,
      lower whisker=39.00
    }, draw=black, fill=LightYellow,
    ] coordinates {};
    \addplot+[
    boxplot prepared={
      median=55.39,
      upper quartile=61.15,
      lower quartile=42.42,
      upper whisker=68.76,
      lower whisker=33.75
    }, draw=black, fill=LightGreen,
    ] coordinates {};
    \addplot+[
    boxplot prepared={
      median=56.96,
      upper quartile=63.85,
      lower quartile=43.31,
      upper whisker=69.97,
      lower whisker=28.73
    }, draw=black, fill=Purple,
    ] coordinates {};
  \end{axis}
\end{tikzpicture}
        \caption{$\tau=20$}
        \label{fig:bw20s}
    \end{subfigure}
    \hspace{-10pt}
    \begin{subfigure}[b]{0.2\textwidth}
        \centering
\begin{tikzpicture}
  \begin{axis}
    [
    boxplot/draw direction=y,
    xmajorgrids=true,
    width= 1.0\linewidth,
    trim left = 5cm,
    xtick={1,2},
    xlabel = $t$,
    xticklabels={0, 1},
    boxplot/box extend=0.35,
    height = 1.0\linewidth,
    ]
    \addplot+[
    boxplot prepared={
      median=54.76,
      upper quartile=59.00,
      lower quartile=51.35,
      upper whisker=65.66,
      lower whisker=44.30
    }, draw=black, fill=LightYellow,
    ] coordinates {};
    \addplot+[
    boxplot prepared={
      median=58.71,
      upper quartile=65.04,
      lower quartile=53.90,
      upper whisker=72.08,
      lower whisker=43.22
    }, draw=black, fill=Purple,
    ] coordinates {};
  \end{axis}
\end{tikzpicture}
        \caption{$\tau=30$}
        \label{fig:bw30s}
    \end{subfigure}
    \hspace{-10pt}
    \begin{subfigure}[b]{0.2\textwidth}
        \centering
\begin{tikzpicture}
  \begin{axis}
    [
    boxplot/draw direction=y,
    xmajorgrids=true,
    width= \linewidth,
    trim left = 10cm,
    xtick={1, 2, 3},
    xlabel = $t$,
    xticklabels={0, 1, 2},
    boxplot/box extend=0.5,
    height = 1.0\linewidth,
    ]
    \addplot+[
    boxplot prepared={
      median=42.75,
      upper quartile=43.48,
      lower quartile=41.57,
      upper whisker=46.36,
      lower whisker=38.87
    }, draw=black, fill=LightYellow,
    ] coordinates {};
    \addplot+[
    boxplot prepared={
      median=47.91,
      upper quartile=49.48,
      lower quartile=45.15,
      upper whisker=54.26,
      lower whisker=38.42
    }, draw=black, fill=LightGreen,
    ] coordinates {};
    \addplot+[
    boxplot prepared={
      median=51.14,
      upper quartile=52.82,
      lower quartile=47.44,
      upper whisker=60.73,
      lower whisker=37.88
    }, draw=black, fill=Purple,
    ] coordinates {};
  \end{axis}
\end{tikzpicture}
        \caption{$\tau=20$ ~(w/ memory)}
        \label{fig:bw20c}
    \end{subfigure}
    \hspace{-10pt}
    \begin{subfigure}[b]{0.2\textwidth}
        \centering
\begin{tikzpicture}
  \begin{axis}
    [
    boxplot/draw direction=y,
    xmajorgrids=true,
    width= 1.0\linewidth,
    trim left = 10cm,
    xtick={1,2},
    xlabel = $t$,
    xticklabels={0, 1},
    boxplot/box extend=0.35,
    height = 1.0\linewidth,
    ]
    \addplot+[
    boxplot prepared={
      median=54.17,
      upper quartile=57.72,
      lower quartile=50.30,
      upper whisker=64.36,
      lower whisker=44.25
    }, draw=black, fill=LightYellow,
    ] coordinates {};
    \addplot+[
    boxplot prepared={
      median=62.19,
      upper quartile=67.00,
      lower quartile=58.39,
      upper whisker=73.20,
      lower whisker=47.57
    }, draw=black, fill=Purple,
    ] coordinates {};
  \end{axis}
\end{tikzpicture}
        \caption{$\tau=30$ ~(w/ memory)}
        \label{fig:bw30c}
    \end{subfigure}
    \caption{Distribution of accuracies achieved on DukeMTMC \cite{DukeMTMC} with R$^2$MMT.}
    \label{fig:stats}
\end{figure*}

One hour of DukeMTMC-video is used as the data stream, split into equal sized continuous segments of size $\tau$. SDS is performed at each time segment on each camera individually, and $k$ refers to the total number of person crops across all training subsets for the full hour. Two methods are used to determine the number of crops needed at each time segment. In the standard method, only data collected in a time segment may be used for training related to that time segment. The second method uses a form of memory, allowing the use of data from the current time segment and previous time segments still in memory. For these experiments, we assume a memory length of up to 60 minutes. \equref{eq:separate} and \equref{eq:combined} are used to calculate the number of person crops needed from each camera at each time segment, for the standard and memory based methods respectively.

\begin{equation} \label{eq:separate}
    k = \sum_{t=0}^{\frac{60}{\tau}-1}\sum_{i=1}^{8}P(C_{i})P(C_{i} \cap \tau_{t})
\end{equation}

\begin{equation} \label{eq:combined}
    k = \sum_{t=0}^{\frac{60}{\tau}-1}\sum_{i=1}^{8}P(C_{i})\sum_{\eta=0}^{t}P(C_{i} \cap \tau_{\eta})
\end{equation}

where $k$ is the total number of person crops desired for the training subset over an hour of video stream, $\tau_{t}$ is a time segment of length $\tau$ minutes that begins at $\tau \times t$ minutes, $C_{i}$ is the $i^{th}$ camera, $P(C_{i})$ is the percentage of total person crops received from $C_{I}$ when compared to all cameras over an hour of video, and $P(C_{i})P(C_{i} \cap \tau_{t})$ is the percentage of person crops received during $\tau_{t}$ for $C_{i}$ compared to all person crops received from $C_{i}$ over an hour of video.

This ensures the number of person crops selected for a subset from each camera at each time segment is proportional to the number of person crops received. 
The variable ranges used in these experiments are shown below.

\begin{equation} \label{eq:phase3vars}
\begin{split}
    K & \in [18, 20, 25, 30, 40, 50] \\
    I & \in [100, 250, 500, 750, 1000, 1500] \\
    E & \in [1, 2, 3, 5] \\
    \tau & \in [15, 20, 30] \\
    t & \in \mathbb{Z} : \{0 \leq t \leq (\frac{60}{\tau}-1)\}
\end{split}
\end{equation}

This creates over 2500 data points across the two methods, becoming difficult to visualize even in three dimensional space. \figref{fig:stats} displays the distribution of training accuracies for each $\tau$ at each time segment. Out of the 864 configurations tested, more than half of them failed to consistently meet the time requirement of R$^2$OUDA and are not included in the statistics. Most notably, all configurations that used memory failed to consistently meet the time requirement when given a $\tau$ of 15. When memory is utilized, the time required for SDS greatly increases for successive time segments as more images accumulate. This limits how large $k$ can be, restricting $K$ to 20 or below when $\tau=20$ and 30 or below when $\tau=30$. Even without memory, the time constraint proves very limiting. Only when $\tau=20$ is the entire range of $K$ able to be utilized. For a more fine grain look at all 2500+ data points in this experiment, please see the supplementary materials.

\begin{table}[t]
    \centering
    \resizebox{0.95\linewidth}{!}{
    \begin{tabular}{c|c||c|c|c|c|c}
            $\tau$ & $t$ & Min & $Q_1$ & $Q_2$ & $Q_3$ & Max \\ 
            \hline \hline
            \rowcolor{DarkGray}
            \multicolumn{7}{c}{R$^2$MMT} \\
            15  & 0 & 35.28 & 40.93 & 43.76 & 47.80 & \cellcolor{LightGray}52.29   \\
                & 1 & 35.68 & 41.43 & 44.48 & 51.66 & \cellcolor{LightGray}57.05   \\
                & 2 & 30.92 & 37.75 & 41.97 & 52.74 & \cellcolor{LightGray}59.92   \\
                & 3 & 26.30 & 33.62 & 40.89 & 51.71 & \cellcolor{LightGray}58.08   \\ \hline
            20  & 0 & 39.00 & 44.39 & 50.27 & 54.29 & \cellcolor{LightGray}61.63   \\
                & 1 & 33.75 & 42.42 & 55.39 & 61.15 & \cellcolor{LightGray}68.76   \\
                & 2 & 28.73 & 43.31 & 56.96 & 63.85 & \cellcolor{LightGray}69.97   \\ \hline
            30  & 0 & 44.30 & 51.35 & 54.76 & 59.04 & \cellcolor{LightGray}65.66   \\
                & 1 & 43.22 & 53.91 & 58.71 & 65.04 & \cellcolor{LightGray}72.08   \\
            \hline \hline
            \rowcolor{DarkGray}
            \multicolumn{7}{c}{R$^2$MMT with memory} \\
            20  & 0 & 38.87 & 41.67 & 42.77 & 43.65 & \cellcolor{LightGray}46.36   \\
                & 1 & 38.42 & 45.20 & 48.03 & 49.87 & \cellcolor{LightGray}54.26   \\
                & 2 & 37.88 & 47.44 & 51.35 & 53.90 & \cellcolor{LightGray}60.73   \\ \hline
            30  & 0 & 44.26 & 50.30 & 54.17 & 57.72 & \cellcolor{LightGray}64.36   \\
                & 1 & 47.58 & 58.39 & 62.17 & 67.00 & \cellcolor{LightGray}73.21   \\
    \end{tabular}
    }
    \caption{Distribution of accuracies achieved on DukeMTMC \cite{DukeMTMC} with R$^2$MMT.}
    \label{tab:boxnwisk}
\end{table}

The data in general follows similar trends as seen in \secref{sec:phase1} and \secref{sec:phase2}, but to more of an extreme. In addition to disqualifying several configurations off the bat, the segmented data stream and time constraint generally mean R$^2$MMT has less data to work with during any given training. Unlike in the previous experiments, the time constraint prevents the system from just throwing more data and more training at the problem. Instead, a balance must be found. We see an overall increase in top accuracies when $\tau$ increases, both in standard and memory configurations. Top accuracies also increase over time, with one notable exception. When $\tau=15$, accuracy actually drops in the final time segment. This is due to the extremely low amount of data available in that particular time segment.

Another interesting observation can be made by looking at $\tau=20$ both with and without memory. While the standard R$^2$MMT achieves higher overall top accuracies, the distribution is a lot more varied when compared to R$^2$MMT with memory. Many configurations actually lose accuracy, far more than when memory is present. This suggests that while memory is limiting, it may add stability to training over time. This is further demonstrated when $\tau=30$. When memory is used the maximum accuracy is lower in the first time segment, being restricted to a lower value of $K$, but is higher in the second time segment due to the increased range of available data. 

\begin{figure}[ht!]
    \centering
    \begin{tikzpicture}
          \begin{axis}[
            legend style = {font=\scriptsize},
            legend pos=north west,
            height=0.6\linewidth,
            xmajorgrids=true,
            ymajorgrids=true,
            xlabel= Time,
            ylabel= Top-1 (\%),
            xmin=0, xmax=60,
            xtick={15, 20, 30, 40, 45, 60},
            xticklabels={15, 20, 30, 40, 45, 60},   
                    ]
        \addplot[mark=*, PastelGreen, style=densely dashed] 
            plot coordinates {
                (0, 42.00)
                (15,52.29)
                (30,57.05)
                (45,58.40)
                (60, 58.08)
        };
        \addlegendentry{S-15}
        
        \addplot[mark=*, color=PastelBlue, style=densely dashed]
            plot coordinates {
                (0, 42.00)
                (20,61.62)
                (40,68.76)
                (60,69.97)
            };
        \addlegendentry{S-20}
        
        \addplot[mark=*, color=PastelPurple, style=densely dashed]
            plot coordinates {
                (0, 42.00)
                (30,64.99)
                (60,72.08)
            };
        \addlegendentry{S-30}
        
        \addplot[mark=*, color=PastelBlue]
            plot coordinates {
                (0, 42.00)
                (20,46.36)
                (40,54.26)
                (60,60.73)
            };
        \addlegendentry{WM-20}
        
        \addplot[mark=*, color=PastelPurple]
            plot coordinates {
                (0, 42.00)
                (30,64.36)
                (60,73.20)
            };
        \addlegendentry{WM-30}
        
        
        
        \legend{} 
        
        \end{axis}
    \end{tikzpicture}
    \caption{Best results for each system configuration. Dashed lines (-~-) represent standard configurations. Solid lines (--) represent configurations with memory. \textcolor{PastelGreen}{Green}, \textcolor{PastelBlue}{blue}, and \textcolor{PastelPurple}{purple} denote $\tau$ values of \textcolor{PastelGreen}{15}, \textcolor{PastelBlue}{20}, and \textcolor{Purple}{30} respectively.}
    \label{fig:phase3_best}
\end{figure}
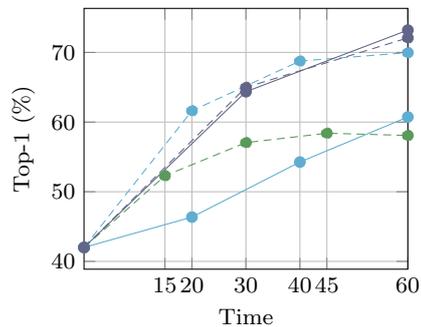

\figref{fig:phase3_best} shows the best configurations of R$^2$MMT, both with and without memory, for each $\tau$. The overall highest accuracy is achieved with memory when $\tau=30$, $K=30$, $E=5$, and $I=500$, reaching an impressive $73.2\%$ Top-1. Despite the much harsher requirements of the R$^2$OUDA setting, this is within $0.1\%$ of the best possible accuracy using MMT in the OUDA setting \cite{OUDA}. However, with a $\tau$ of 30 it also has a latency of 60 minutes between collecting data and inferencing with a model trained on that data. This can be reduced to 30 minutes by changing $\tau$ to 15, but then accuracy drops to a disappointing $58.08\%$. A $\tau$ of 20 splits the difference, achieving a final Top-1 of $69.97\%$ while reducing the inference latency to 40 minutes. This is within $4\%$ of our best overall result, and reduces the delay by over $30\%$.

The strict time constraint disqualified many of the configurations in \secref{sec:phase3}. However, if we ignore the time constraint for a moment we see accuracies reaching up to $76.53\%$ when $\tau=15$, $K=40$, $E=5$, and $I=1500$ in a system with memory, putting it within $1.5\%$ of MMT in the UDA setting \cite{MMT}. With further optimization or more powerful hardware, R$^2$MMT might be able to achieve higher accuracies with decreased latency between collection and inference. This shows that there is a lot of room for improvement and growth in the R$^2$OUDA setting. The explorations in this paper can serve as a guideline for future works. 



\section{Conclusion} \label{sec:conclusion}
This paper proposed the setting of R$^2$OUDA, to better represent the unique challenges of real-world applications. R$^2$MMT was introduced as the first attempt at a real-world, end-to-end system that can address all the demands of the R$^2$OUDA setting. An exhaustive set of experiments were conducted, using R$^2$MMT to create over 100 data subsets and train more than 3000 models, exploring the breadth of the R$^2$OUDA setting. While meeting the harsh requirements of R$^2$OUDA, R$^2$MMT was able to achieve over $73\%$ Top-1 accuracy, reaching within $0.1\%$ of comparable SotA OUDA approaches that cannot be directly applied to real-world applications.


\begin{acknowledgements}
This research is supported by the National Science Foundation (NSF) under Award No. 1831795 and NSF Graduate Research Fellowship Award No. 1848727.
\end{acknowledgements}

%



\bibliographystyle{spmpsci}
\bibliography{egbib}

\begin{thebibliography}{10}
\providecommand{\url}[1]{{#1}}
\providecommand{\urlprefix}{URL }
\expandafter\ifx\csname urlstyle\endcsname\relax
  \providecommand{\doi}[1]{DOI~\discretionary{}{}{}#1}\else
  \providecommand{\doi}{DOI~\discretionary{}{}{}\begingroup
  \urlstyle{rm}\Url}\fi

\bibitem{kmeans++}
Arthur, D., Vassilvitskii, S.: K-means++: The advantages of careful seeding.
\newblock In: Proceedings of the Eighteenth Annual ACM-SIAM Symposium on
  Discrete Algorithms, SODA '07, p. 1027–1035. Society for Industrial and
  Applied Mathematics, USA (2007)

\bibitem{InstanceGuided}
Chen, Y., Zhu, X., Gong, S.: Instance-guided context rendering for cross-domain
  person re-identification.
\newblock In: 2019 IEEE/CVF International Conference on Computer Vision (ICCV),
  pp. 232--242 (2019).
\newblock \doi{10.1109/ICCV.2019.00032}

\bibitem{CANUReID}
Delorme, G., Xu, Y., Lathuiliere, S., Horaud, R., Alameda-Pineda, X.:
  Canu-reid: A conditional adversarial network for unsupervised person
  re-identification.
\newblock In: 2020 25th International Conference on Pattern Recognition (ICPR),
  pp. 4428--4435. IEEE Computer Society, Los Alamitos, CA, USA (2021).
\newblock \doi{10.1109/ICPR48806.2021.9412431}.
\newblock
  \urlprefix\url{https://doi.ieeecomputersociety.org/10.1109/ICPR48806.2021.9412431}

\bibitem{PreservedSelf}
Deng, W., Zheng, L., Ye, Q., Kang, G., Yang, Y., Jiao, J.: Image-image domain
  adaptation with preserved self-similarity and domain-dissimilarity for person
  re-identification.
\newblock In: Proceedings of the IEEE Conference on Computer Vision and Pattern
  Recognition (CVPR) (2018)

\bibitem{DBSCAN}
Ester, M., Kriegel, H.P., Sander, J., Xu, X., et~al.: A density-based algorithm
  for discovering clusters in large spatial databases with noise.
\newblock In: kdd, vol.~96, pp. 226--231 (1996)

\bibitem{OUDACovid19}
Ewen, N., Khan, N.: Online unsupervised learning for domain shift in covid-19
  ct scan datasets.
\newblock In: 2021 IEEE International Conference on Autonomous Systems (ICAS),
  pp. 1--5 (2021).
\newblock \doi{10.1109/ICAS49788.2021.9551146}

\bibitem{ClusteringFineTuning}
Fan, H., Zheng, L., Yan, C., Yang, Y.: Unsupervised person re-identification:
  Clustering and fine-tuning.
\newblock ACM Trans. Multimedia Comput. Commun. Appl. \textbf{14}(4) (2018).
\newblock \doi{10.1145/3243316}.
\newblock \urlprefix\url{https://doi.org/10.1145/3243316}

\bibitem{ComplementaryPseudoLabels}
Feng, H., Chen, M., Hu, J., Shen, D., Liu, H., Cai, D.: Complementary pseudo
  labels for unsupervised domain adaptation on person re-identification.
\newblock IEEE Transactions on Image Processing \textbf{30}, 2898--2907 (2021).
\newblock \doi{10.1109/TIP.2021.3056212}

\bibitem{MemoryConstraints}
Fini, E., Lathuili{\`e}re, S., Sangineto, E., Nabi, M., Ricci, E.: Online
  continual learning under extreme memory constraints.
\newblock In: A.~Vedaldi, H.~Bischof, T.~Brox, J.M. Frahm (eds.) Computer
  Vision -- ECCV 2020, pp. 720--735. Springer International Publishing, Cham
  (2020)

\bibitem{SelfSimilarityGrouping}
Fu, Y., Wei, Y., Wang, G., Zhou, Y., Shi, H., Huang, T.S.: Self-similarity
  grouping: A simple unsupervised cross domain adaptation approach for person
  re-identification.
\newblock In: Proceedings of the IEEE/CVF International Conference on Computer
  Vision (ICCV) (2019)

\bibitem{MMT}
Ge, Y., Chen, D., Li, H.: Mutual mean-teaching: Pseudo label refinery for
  unsupervised domain adaptation on person re-identification.
\newblock In: International Conference on Learning Representations (2020).
\newblock \urlprefix\url{https://openreview.net/forum?id=rJlnOhVYPS}

\bibitem{SpCL}
Ge, Y., Zhu, F., Chen, D., Zhao, R., Li, H.: Self-paced contrastive learning
  with hybrid memory for domain adaptive object re-id.
\newblock In: Advances in Neural Information Processing Systems (2020)

\bibitem{SelfPacedContrast}
Ge, Y., Zhu, F., Chen, D., Zhao, R., Li, H.: Self-paced contrastive learning
  with hybrid memory for domain adaptive object re-id.
\newblock In: Advances in Neural Information Processing Systems (2020)

\bibitem{RelationRegularization}
Ge, Y., Zhu, F., Chen, D., Zhao, R., Wang, X., Li, H.: Structured domain
  adaptation with online relation regularization for unsupervised person re-id
  (2020).
\newblock \doi{10.48550/ARXIV.2003.06650}.
\newblock \urlprefix\url{https://arxiv.org/abs/2003.06650}

\bibitem{GANs}
Goodfellow, I., Pouget-Abadie, J., Mirza, M., Xu, B., Warde-Farley, D., Ozair,
  S., Courville, A., Bengio, Y.: Generative adversarial nets.
\newblock In: Z.~Ghahramani, M.~Welling, C.~Cortes, N.~Lawrence, K.~Weinberger
  (eds.) Advances in Neural Information Processing Systems, vol.~27. Curran
  Associates, Inc. (2014).
\newblock
  \urlprefix\url{https://proceedings.neurips.cc/paper/2014/file/5ca3e9b122f61f8f06494c97b1afccf3-Paper.pdf}

\bibitem{ResNet}
He, K., Zhang, X., Ren, S., Sun, J.: Deep residual learning for image
  recognition.
\newblock 2016 IEEE Conference on Computer Vision and Pattern Recognition
  (CVPR) pp. 770--778 (2016)

\bibitem{OUDAECG}
He, W., Ye, Y., Li, Y., Pan, T., Lu, L.: Online cross-subject emotion
  recognition from ecg via unsupervised domain adaptation.
\newblock In: 2021 43rd Annual International Conference of the IEEE Engineering
  in Medicine \& Biology Society (EMBC), pp. 1001--1005 (2021).
\newblock \doi{10.1109/EMBC46164.2021.9630433}

\bibitem{DefenseTripletLoss}
Hermans*, A., Beyer*, L., Leibe, B.: {In Defense of the Triplet Loss for Person
  Re-Identification}.
\newblock arXiv preprint arXiv:1703.07737  (2017)

\bibitem{SBSGAN}
Huang, Y., Wu, Q., Xu, J., Zhong, Y.: Sbsgan: Suppression of inter-domain
  background shift for person re-identification.
\newblock In: 2019 IEEE/CVF International Conference on Computer Vision (ICCV),
  pp. 9526--9535. IEEE Computer Society, Los Alamitos, CA, USA (2019).
\newblock \doi{10.1109/ICCV.2019.00962}.
\newblock
  \urlprefix\url{https://doi.ieeecomputersociety.org/10.1109/ICCV.2019.00962}

\bibitem{Pix2Pix}
Isola, P., Zhu, J.Y., Zhou, T., Efros, A.: Image-to-image translation with
  conditional adversarial networks.
\newblock In: 2017 IEEE/CVF Conference on Computer Vision and Pattern
  Recognition (CVPR), pp. 5967--5976 (2017).
\newblock \doi{10.1109/CVPR.2017.632}

\bibitem{GreedyKCenters}
Jali, N., Karamchandani, N., Moharir, S.: Greedy kk-center from noisy distance
  samples.
\newblock IEEE Transactions on Signal and Information Processing over Networks
  \textbf{8}, 330--343 (2022).
\newblock \doi{10.1109/TSIPN.2022.3164352}

\bibitem{YOLOv5}
Jocher, G., Chaurasia, A., Stoken, A., Borovec, J., NanoCode012, Kwon, Y.,
  TaoXie, Michael, K., Fang, J., imyhxy, Lorna, Wong, C., Yifu, Z., V, A.,
  Montes, D., Wang, Z., Fati, C., Nadar, J., Laughing, UnglvKitDe, tkianai,
  yxNONG, Skalski, P., Hogan, A., Strobel, M., Jain, M., Mammana, L., xylieong:
  {ultralytics/yolov5: v6.2 - YOLOv5 Classification Models, Apple M1,
  Reproducibility, ClearML and Deci.ai integrations} (2022).
\newblock \doi{10.5281/zenodo.7002879}.
\newblock \urlprefix\url{https://doi.org/10.5281/zenodo.7002879}

\bibitem{OUDASeg}
Kuznietsov, Y., Proesmans, M., Gool, L.V.: Towards unsupervised online domain
  adaptation for semantic segmentation.
\newblock In: 2022 IEEE/CVF Winter Conference on Applications of Computer
  Vision Workshops (WACVW), pp. 261--271 (2022).
\newblock \doi{10.1109/WACVW54805.2022.00032}

\bibitem{OpenWorldSurvey}
Leng, Q., Ye, M., Tian, Q.: A survey of open-world person re-identification.
\newblock IEEE Transactions on Circuits and Systems for Video Technology
  \textbf{30}(4), 1092--1108 (2020).
\newblock \doi{10.1109/TCSVT.2019.2898940}

\bibitem{CUHK03}
Li, W., Zhao, R., Xiao, T., Wang, X.: Deepreid: Deep filter pairing neural
  network for person re-identification.
\newblock In: CVPR (2014)

\bibitem{COCO}
Lin, T.Y., Maire, M., Belongie, S., Hays, J., Perona, P., Ramanan, D.,
  Doll{\'a}r, P., Zitnick, C.L.: Microsoft coco: Common objects in context.
\newblock In: D.~Fleet, T.~Pajdla, B.~Schiele, T.~Tuytelaars (eds.) Computer
  Vision -- ECCV 2014, pp. 740--755. Springer International Publishing, Cham
  (2014)

\bibitem{BottumUpClustering}
Lin, Y., Dong, X., Zheng, L., Yan, Y., Yang, Y.: A bottom-up clustering
  approach to unsupervised person re-identification.
\newblock Proceedings of the AAAI Conference on Artificial Intelligence
  \textbf{33}(01), 8738--8745 (2019).
\newblock \doi{10.1609/aaai.v33i01.33018738}.
\newblock \urlprefix\url{https://ojs.aaai.org/index.php/AAAI/article/view/4898}

\bibitem{AdaptiveTransfer}
Liu, J., Zha, Z.J., Chen, D., Hong, R., Wang, M.: Adaptive transfer network for
  cross-domain person re-identification.
\newblock In: 2019 IEEE/CVF Conference on Computer Vision and Pattern
  Recognition (CVPR), pp. 7195--7204 (2019).
\newblock \doi{10.1109/CVPR.2019.00737}

\bibitem{OUDAMultistage}
Moon, J., Das, D., George~Lee, C.S.: A multistage framework with mean subspace
  computation and recursive feedback for online unsupervised domain adaptation.
\newblock IEEE Transactions on Image Processing \textbf{31}, 4622--4636 (2022).
\newblock \doi{10.1109/TIP.2022.3186537}

\bibitem{OUDAMultistep}
Moon, J.H., Das, D., Lee, C.G.: Multi-step online unsupervised domain
  adaptation.
\newblock In: ICASSP 2020 - 2020 IEEE International Conference on Acoustics,
  Speech and Signal Processing (ICASSP), pp. 41172--41576 (2020).
\newblock \doi{10.1109/ICASSP40776.2020.9052976}

\bibitem{REVAMP2T}
Neff, C., Mendieta, M., Mohan, S., Baharani, M., Rogers, S., Tabkhi, H.:
  Revamp2t: Real-time edge video analytics for multicamera privacy-aware
  pedestrian tracking.
\newblock IEEE Internet of Things Journal \textbf{7}(4), 2591--2602 (2020).
\newblock \doi{10.1109/JIOT.2019.2954804}

\bibitem{flipreid}
Ni, X., Rahtu, E.: Flipreid: Closing the gap between training and inference in
  person re-identification.
\newblock In: 2021 9th European Workshop on Visual Information Processing
  (EUVIP), pp. 1--6 (2021).
\newblock \doi{10.1109/EUVIP50544.2021.9484010}

\bibitem{OUDA}
Rami, H., Ospici, M., Lathuili\`ere, S.: Online unsupervised domain adaptation
  for person re-identification.
\newblock In: Proceedings of the IEEE/CVF Conference on Computer Vision and
  Pattern Recognition (CVPR) Workshops, pp. 3830--3839 (2022)

\bibitem{HeuristicKCenters}
Rana, R., Garg, D.: Heuristic approaches for k-center problem.
\newblock In: 2009 IEEE International Advance Computing Conference, pp.
  332--335 (2009).
\newblock \doi{10.1109/IADCC.2009.4809031}

\bibitem{DukeMTMC}
Ristani, E., Solera, F., Zou, R., Cucchiara, R., Tomasi, C.: Performance
  measures and a data set for multi-target, multi-camera tracking.
\newblock In: European Conference on Computer Vision workshop on Benchmarking
  Multi-Target Tracking (2016)

\bibitem{TheoryPractice}
Song, L., Wang, C., Zhang, L., Du, B., Zhang, Q., Huang, C., Wang, X.:
  Unsupervised domain adaptive re-identification: Theory and practice.
\newblock Pattern Recogn. \textbf{102}(C) (2020).
\newblock \doi{10.1016/j.patcog.2019.107173}.
\newblock \urlprefix\url{https://doi.org/10.1016/j.patcog.2019.107173}

\bibitem{HRNet}
Sun, K., Xiao, B., Liu, D., Wang, J.: Deep high-resolution representation
  learning for human pose estimation.
\newblock In: CVPR (2019)

\bibitem{MeanTeachers}
Tarvainen, A., Valpola, H.: Mean teachers are better role models:
  Weight-averaged consistency targets improve semi-supervised deep learning
  results.
\newblock In: Proceedings of the 31st International Conference on Neural
  Information Processing Systems, NIPS'17, p. 1195–1204. Curran Associates
  Inc., Red Hook, NY, USA (2017)

\bibitem{OUDASegFrequency}
Termöhlen, J.A., Klingner, M., Brettin, L.J., Schmidt, N.M., Fingscheidt, T.:
  Continual unsupervised domain adaptation for semantic segmentation by online
  frequency domain style transfer.
\newblock In: 2021 IEEE International Intelligent Transportation Systems
  Conference (ITSC), pp. 2881--2888 (2021).
\newblock \doi{10.1109/ITSC48978.2021.9564566}

\bibitem{STReID}
Wang, G., Lai, J., Huang, P., Xie, X.: Spatial-temporal person
  re-identification.
\newblock Proceedings of the AAAI Conference on Artificial Intelligence
  \textbf{33}(01), 8933--8940 (2019).
\newblock \doi{10.1609/aaai.v33i01.33018933}.
\newblock \urlprefix\url{https://ojs.aaai.org/index.php/AAAI/article/view/4921}

\bibitem{CascadedPairwise}
Wang, Y., Chen, Z., Wu, F., Wang, G.: Person re-identification with cascaded
  pairwise convolutions.
\newblock In: 2018 IEEE/CVF Conference on Computer Vision and Pattern
  Recognition, pp. 1470--1478 (2018).
\newblock \doi{10.1109/CVPR.2018.00159}

\bibitem{PersonTransferGAN}
Wei, L., Zhang, S., Gao, W., Tian, Q.: Person transfer gan to bridge domain gap
  for person re-identification.
\newblock In: Proceedings of the IEEE Conference on Computer Vision and Pattern
  Recognition (CVPR) (2018)

\bibitem{Centroids}
Wieczorek, M., Rychalska, B., Dabrowski, J.: On the unreasonable effectiveness
  of centroids in image retrieval.
\newblock ArXiv \textbf{abs/2104.13643} (2021)

\bibitem{DeepSORT}
Wojke, N., Bewley, A., Paulus, D.: Simple online and realtime tracking with a
  deep association metric.
\newblock In: 2017 IEEE International Conference on Image Processing (ICIP),
  pp. 3645--3649 (2017).
\newblock \doi{10.1109/ICIP.2017.8296962}

\bibitem{DynamicGraph}
Ye, M., Li, J., Ma, A.J., Zheng, L., Yuen, P.C.: Dynamic graph co-matching for
  unsupervised video-based person re-identification.
\newblock IEEE Transactions on Image Processing \textbf{28}(6), 2976--2990
  (2019).
\newblock \doi{10.1109/TIP.2019.2893066}

\bibitem{Survey2020}
Ye, M., Shen, J., Lin, G., Xiang, T., Shao, L., Hoi, S.C.H.: Deep learning for
  person re-identification: A survey and outlook (2020).
\newblock \doi{10.48550/ARXIV.2001.04193}.
\newblock \urlprefix\url{https://arxiv.org/abs/2001.04193}

\bibitem{OUDADiscrepancies}
Ye, Y., Pan, T., Meng, Q., Li, J., Shen, H.T.: Online unsupervised domain
  adaptation via reducing inter- and intra-domain discrepancies.
\newblock IEEE Transactions on Neural Networks and Learning Systems pp. 1--15
  (2022).
\newblock \doi{10.1109/TNNLS.2022.3177769}

\bibitem{HCR}
Zeng, K.: Hierarchical clustering with hard-batch triplet loss for person
  re-identification (2019).
\newblock \doi{10.48550/ARXIV.1910.12278}.
\newblock \urlprefix\url{https://arxiv.org/abs/1910.12278}

\bibitem{MARS}
Zheng, L., Bie, Z., Sun, Y., Wang, J., Su, C., Wang, S., Tian, Q.: Mars: A
  video benchmark for large-scale person re-identification.
\newblock In: B.~Leibe, J.~Matas, N.~Sebe, M.~Welling (eds.) Computer Vision --
  ECCV 2016, pp. 868--884. Springer International Publishing, Cham (2016)

\bibitem{Market}
Zheng, L., Shen, L., Tian, L., Wang, S., Wang, J., Tian, Q.: Scalable person
  re-identification: A benchmark.
\newblock In: 2015 IEEE International Conference on Computer Vision (ICCV), pp.
  1116--1124 (2015).
\newblock \doi{10.1109/ICCV.2015.133}

\bibitem{PastPresentFuture}
Zheng, L., Yang, Y., Hauptmann, A.: Person re-identification: Past, present and
  future.
\newblock ArXiv \textbf{abs/1610.02984} (2016)

\bibitem{ReIDWild}
Zheng, L., Zhang, H., Sun, S., Chandraker, M., Yang, Y., Tian, Q.: Person
  re-identification in the wild.
\newblock In: 2017 IEEE Conference on Computer Vision and Pattern Recognition
  (CVPR), pp. 3346--3355 (2017).
\newblock \doi{10.1109/CVPR.2017.357}

\bibitem{kreciprocal}
Zhong, Z., Zheng, L., Cao, D., Li, S.: Re-ranking person re-identification with
  k-reciprocal encoding.
\newblock In: Proceedings of the IEEE Conference on Computer Vision and Pattern
  Recognition (CVPR) (2017)

\bibitem{UnpairedI2I}
Zhu, J.Y., Park, T., Isola, P., Efros, A.A.: Unpaired image-to-image
  translation using cycle-consistent adversarial networks.
\newblock In: 2017 IEEE International Conference on Computer Vision (ICCV), pp.
  2242--2251 (2017).
\newblock \doi{10.1109/ICCV.2017.244}

\bibitem{VALossReID}
Zhu, Z., Jiang, X., Zheng, F., Guo, X., Huang, F., Zheng, W., Sun, X.:
  Viewpoint-aware loss with angular regularization for person re-identification
  (2019).
\newblock \doi{10.48550/ARXIV.1912.01300}.
\newblock \urlprefix\url{https://arxiv.org/abs/1912.01300}

\end{thebibliography}

\end{document}